\documentclass[letterpaper, 10 pt, conference]{ieeeconf}  

\IEEEoverridecommandlockouts                              
\usepackage{amsmath}
\usepackage{amssymb}
\usepackage{color, colortbl}
\usepackage[dvipsnames]{xcolor}
\usepackage{bm}
\usepackage{bbm}
\usepackage{comment}
\usepackage{tikz}
\usepackage{cite}
\usepackage{dsfont}
\usetikzlibrary{shapes,arrows,positioning,graphs,decorations.pathreplacing,angles,quotes,arrows.meta,calc,automata,matrix}
\usepackage{tabularx}
\usetikzlibrary{external, shapes,arrows, positioning, matrix, calc}

\tikzset{%
  block/.style    = {draw, rectangle, minimum height = 5em,
    text width = 5em, text centered, rounded corners=1.5mm, node distance = 3cm, fill=white},
  sum/.style      = {draw, circle, node distance = 1cm}, 
  input/.style    = {coordinate}, 
  output/.style   = {coordinate} 
}

\newtheorem{remark}{Remark}

\title{\LARGE \bf
Optimal Transport for Correctional Learning
}

\author{Rebecka Winqvist, In\^{e}s Louren\c{c}o, Francesco Quinzan, Cristian R.\ Rojas and Bo Wahlberg
\thanks{This work was supported by the Wallenberg AI, Autonomous Systems and Software Program
        (WASP), the Swedish Research Council Research Environment NewLEADS under contract 2016-06079, and the Digital Futures project EXTREMUM.}
\thanks{The authors are with the Division of Decision and Control Systems, KTH Royal Institute of Technology, Stockholm, Sweden 
        {\tt\small \{rebwin, ineslo, quinzan, crro, bo\}@kth.se }}
        }
\begin{document}

\maketitle
\thispagestyle{empty}
\pagestyle{empty}

\begin{abstract}
    The contribution of this paper is a generalized formulation of correctional learning using optimal transport, which is about how to optimally transport one mass distribution to another.
    Correctional learning is a framework developed to enhance the accuracy of parameter estimation processes by means of a teacher-student approach. In this framework, an expert agent, referred to as the teacher, 
    modifies the data used by a learning agent, known as the student, to improve its estimation process. The objective of the teacher is to alter the data such that the student's estimation error is minimized, subject to a 
    fixed intervention budget. Compared to existing formulations of correctional learning, our novel optimal transport approach provides several benefits. It allows for the estimation of more complex characteristics as well as the consideration of multiple intervention policies for the teacher. We evaluate our approach on two theoretical examples, and on a human-robot interaction application in which the teacher's role is to improve the robots performance in an inverse reinforcement learning setting.
\end{abstract}

\section{Introduction}
Parameter estimation refers to the process of determining a model's parameter values based on measured data. The values of these parameters have a direct impact on the distribution of the data generated by the modeled system. Estimation theory is a well-researched topic with several established methods, see e.g.\ \cite{Ljung1999}. The interest in the subject is widespread with applications to be found in, among others, the process industries, control applications, as well as in the research of biological functions and systems \cite{Astrom1971Survey}. Popular estimation methods range from the conventional maximum likelihood and Bayesian inference methods, to more recent learning-based methods. 

Common to most of these estimators is their data-driven nature. In many real-world applications, however, the available data often does not accurately reflect the underlying distribution or behavior of the system being studied. This can be the result of limited sample sizes, biased sampling methods, measurement errors, or outliers \cite{Liu2022}. Relying on such data when using data-driven estimators can result in inaccurate parameter estimation and poor model performance. For example, a recent study found that commercially available facial analysis algorithms showed higher error rates for darker-skinned individuals and women \cite{pmlr-v81-buolamwini18a}, which could be linked to unrepresentative training data. 

\textit{Correctional learning} is a recently developed framework that may be used to address this issue \cite{lourencco2021cooperative, Lourenco2022}. The framework arises from the idea of cooperative (learning) problems, i.e., settings in which two or more agents work together towards a common goal. The framework is structured around a teacher-student model, where an expert (teacher) agent seeks to assist a learner (student) agent in its estimation process. More specifically, the teacher's goal is to modify (or correct) the collected observations, based on which the student forms its estimation. See Figure \ref{fig:correctional_learning} for an illustration of this. Correctional learning can thus be viewed as a means of finding an optimal mapping from the original observations to a modified sequence that minimizes the student's estimation error.

Recent works on correctional learning have shown promising results in both offline and online settings \cite{lourencco2021cooperative, Lourenco2022}. Nevertheless, the framework still suffers from some limitations. First of all, the derived performance guarantees hold only for simple systems. To describe real-world phenomena, however, one typically requires more complex distributions. For example, a Gaussian distribution can be used to describe biological data such as the heights of people. To model the probability of failure of an appliance, we can use the Weibull distribution. Another disadvantage is that the teacher's policy follows explicitly from the solution, leaving no room for alternative intervention strategies to be considered.

Our contribution is an alternative approach to correctional learning using tools from optimal transport \cite{Villani2003}. Optimal transport is a mathematical framework concerned with finding the most efficient way to transport mass from one location to another, according to some cost function. Historically, optimal transport has been widely used in finance and logistics \cite{Galichon2016}, but recent advances have made it an increasingly popular tool in fields such as systems, control and estimation \cite{Haasler2021, Chen2021}. In machine learning, optimal transport has found use in a number of applications, including shape reconstruction \cite{Digne2014}, multi-label classification \cite{NIPS2015_a9eb8122}, and brain decoding \cite{Gramfort2015}. Moreover, recent work in robotics demonstrates how optimal transport can be applied to mapping problems to enable robots to operate in new environments \cite{Tompkins2020}, and for policy fusion in reinforcement learning to speed up the process of a robot learning a new task \cite{Tan2022}.

In the context of correctional learning, we note that the optimal corrections can be viewed as a transportation of probability mass from an initial distribution into a target distribution. Furthermore, by assuming that the estimator depends on the samples only through their empirical measure, we can pose the correctional learning problem as an optimization program in terms of distribution functions -- i.e., as an optimal transport problem. In contrast to \cite{lourencco2021cooperative} and \cite{Lourenco2022}, this novel formulation considers the samples implicitly through their distribution, which not only enables the estimation of more complex parameters, but also allows for the consideration of alternative intervention strategies.

The main contributions of this paper are:
\begin{itemize}
    \item \textit{A generalized correctional learning framework: } we leverage the principles of optimal transport and propose a novel formulation of correctional learning. With this new framework, we can expand the range of applications to consider more sophisticated tasks that involve complex systems.
    \item \textit{Multiple teacher policies: } in standard learning settings, a teacher agent may exhibit several intervention strategies. We show how our new formulation allows for the consideration of multiple teacher policies to fit different tasks.
    \item \textit{Evaluation of performance: } we demonstrate the benefits of our optimal-transport approach by applying the framework on three different test cases. Specifically, we show how the framework can be used to estimate the parameters of more complex distributions such as the Gaussian and the Weibull. We also apply the framework to update a robot's reward function in an inverse reinforcement learning setting.
\end{itemize}

\subsection{Related Work}
Learning from experts is a widely studied problem in machine learning. Learning from demonstrations \cite{ARGALL2009469} and imitation learning \cite{husseinImitation} are two closely related paradigms where a robot learns from observing the behavior of an expert. In corrective feedback \cite{Najar2021}, on the other hand, the expert provides corrections to the robot's actions to improve its learning process. This is opposed to correctional learning, where the corrections are made to the data that the robot learns from.

By interpreting correctional learning as a means for customizing a data set to better suit a specific learning task, we find similarities with other techniques in machine learning and statistics. Feature selection \cite{KIRA1992249} is one such example, where the aim is to find the most informative and relevant features to improve learning. Other examples include active learning \cite{AggrawalActiveLearning}, where the learner can interactively ask the expert to label new data, and input for system identification \cite{HJALMARSSON2009275, PRONZATO2008303}, where the aim is to design input data to optimize for a model's accuracy. 

In the context of system identification and estimation, our framework can also be placed around other works that also use tools from optimal transport. For example, in \cite{chen2018state}, the authors use optimal transport for state tracking of linear ensembles. More specifically, they propose an optimal-transport approach for estimating the states of multiple subsystems based on their joint output. Other related works include \cite{Taghvaei2021}, where they propose an optimal transport formulation of the ensemble Kalman filter, and \cite{Yang2023}, where the authors study the use of optimal transport distances as objective functions for parameter estimation in dynamical systems.  

\section{Preliminaries}
In this section, we first define the notation used throughout the paper. We then give a brief introduction to correctional learning and optimal transport.

\subsection{Notation}
We use $(\mathcal{Y},\mathcal{B})$ to denote a measurable space, in which $\mathcal{Y}$ is a set, and $\mathcal{B}$ is a $\sigma$-algebra of subsets of $\mathcal{Y}$. We denote the set of positive measures on $(\mathcal{Y},\mathcal{B})$ by $\mathcal{M}_+(\mathcal{Y})$. For $N \in \mathbb{N}$, we define $[N] := \{1,2,\hdots,N\}$. For a set $\mathcal{A}$, we use $|\mathcal{A}|$ to denote its cardinality. Inequalities between vectors and matrices are considered element-wise. We use the words samples and observations interchangeably throughout the paper. 

\subsection{Correctional Learning} \label{subsec:corr_learning}
Consider a model of some data-generating system parameterized by the unknown parameter $\theta \in \Theta$. Let the true system correspond to the value $\theta_0$. In a standard parameter estimation setting, a learner (student) agent aims to estimate $\theta_0$ as $\hat{\theta}_N$, based on a sequence of observations sampled from the system, $\mathcal{O}_N = (x_1, \hdots, x_N)$, distributed according to $p_0^N \in \mathcal{M}_+(\mathcal{X}^N)$, where $x_i \in \mathcal{X} \subseteq \mathbb{R}^d$ for all $i$, and $(\mathcal{X}, \mathcal{B})$ is a measurable space. That is, 
\begin{equation}
    \hat{\theta}_N = f_N(\mathcal{O}_N) = f_N(x_1, \hdots, x_N),
\end{equation}
where $f_N\colon \mathcal{X}^N \to \Theta$ is some estimator function. In the rest of the paper, we will omit the dependence on $N$ if its value is clear from the context.

In the correctional learning framework, an expert (teacher) agent is introduced to help the the student in its estimation process. The teacher may do so by modifying the original observation sequence, $\mathcal{O}$, into a sequence, $\tilde{\mathcal{O}}$, that better represents the true characteristics of the system. The modified sequence is then passed on to the student, who forms the altered estimate $\Tilde{\theta}$.

However, utilizing expert knowledge might be expensive or limited. 
The number of allowed interventions might also be restricted for privacy preserving reasons -- the more observations the teacher changes, the more likely it is to be discovered. To account for this, the teacher is constrained to not exceed a certain intervention budget $B$. If $C_N\colon \mathcal{X}^N \times \mathcal{X}^N \to \mathbb{R}_0^+$ denotes a distance measure between two sequences of $N$ elements, then the teacher must satisfy 
\begin{equation}
    C_N(\mathcal{O}_N,\tilde{\mathcal{O}}_N) \leq B.
\end{equation}
The cost may be chosen to be any distance metric, e.g.\ the $L_1$-norm for discrete observations.

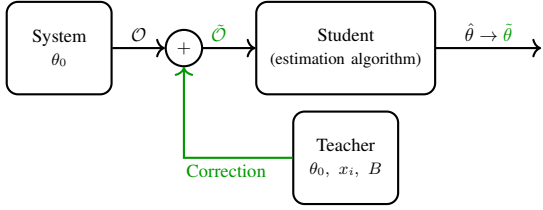
\begin{figure}
    \centering
    \begin{tikzpicture}[auto, node distance=1cm, scale=0.7, transform shape, thick]
        \draw
	node [block] (system) {System  \ \ \ \ \ \ {\small $\theta_0$}}
	node [sum, right=1cm of system, fill=white] (sum) {$+$}
	node [block, right=1cm of sum, text width = 9em] (student) {Student \ \ \ \ \ \ {\small(estimation algorithm)}}
	node [block, below=0.3cm of student] (teacher) {Teacher \ \ \ \ \ \ {\small $\theta_0, \ x_i, \ B$}}	
	node [output, right=2cm of student] (estimate) {};
	
	\draw[->] (system) -- node[above, midway] {$\mathcal{O}$} (sum);
	\draw[->] (sum) -- node[above, pos=0.3]  {{\color{black!40!green}$\Tilde{\mathcal{O}}$}} (student);
	\draw[->] (student) -- node {$\hat{\theta} \rightarrow {\color{black!40!green}\Tilde{\theta}}$} (estimate);
	\draw[->, color=black!40!green] (teacher)  -| node[right=0.8cm, below] {{Correction}} (sum);
    \end{tikzpicture}
    \caption{A schematic view of the correctional learning framework. The teacher knows the true parameter value $\theta_0$ and the original samples $x_i$. The teacher modifies the original sequence of observations, $\mathcal{O}$, into $\tilde{\mathcal{O}}$ by changing at most $B$ samples.}
    \label{fig:correctional_learning}
\end{figure}
The goal of the teacher agent is to find the optimal modified sequence that minimizes the student's estimation error
\begin{equation}
    V(\theta_0, \tilde{\theta}) = \lVert \theta_0 - \tilde{\theta} \rVert.
\end{equation}
where $\| \cdot \|$ is a norm on $\Theta$.

Depending on the setting, this problem can be posed and solved in different ways: 

\subsubsection{Batch setting}
In the offline (batch) setting, the observations are made available to the teacher in batches. By having access to all samples at once, it is shown in \cite{lourencco2021cooperative} that the offline problem can be cast as the optimization program
\begin{equation}
    \begin{aligned}
    \min_{\tilde{\mathcal{O}}  }  \quad & V(\theta_0, \tilde{\theta}) \\
    \text{s.t.} \quad & \tilde{x}_i \in \mathcal{X}, \text{ for all } \tilde{x}_i \in \tilde{\mathcal{O}},\\
     &  C( \mathcal{O} , \tilde{\mathcal{O}} ) \leq B.
    \end{aligned}
\label{eq:cl_batch_opt_prob}
\end{equation}

\subsubsection{Online setting}
In the online setting, the observations are made available sequentially (one at a time). This means that the teacher has to decide, at each time step, whether or not to change the new incoming observation. In \cite{Lourenco2022}, the authors show how the online problem can be formulated as a Markov decision process, and solved using dynamic programming.

 A schematic view of a general correctional learning framework is provided in Figure \ref{fig:correctional_learning}. 

\subsection{Optimal Transport}
Assume that we are given a probability measure $\mu \in \mathcal{M}_+(\mathcal{X})$, which can be interpreted as, say, a distribution of sand in $\mathcal{X}$ of total mass $1$. Assume further that we wish to transform $\mu$ into another probability measure $\nu \in \mathcal{M}_+(\mathcal{Y})$, corresponding to a different distribution of sand, by ``moving'' the grains of sand with minimal transportation cost.
The cost of transporting one unit of probability mass from location $x$ to location $y$ is quantified by a metric on $\mathcal{X}$, $c(x,y)$. To compute the total transportation cost, we must also define a transportation map, which is modelled by a probability measure $\pi \in \mathcal{M}_+(\mathcal{X} \times \mathcal{Y})$, with $d\pi(x,y)$ denoting the amount of mass transferred from $x$ to $y$. Since we cannot move more mass than what we originally have, it must hold that the mass moved from one point in $\mu$ must be received by $\nu$ and vice versa. In mathematical terms, we express those conditions by 
\begin{equation}
    \int_{\mathcal{Y}}d\pi(x,y) = d\mu(x) \quad \text { and }\quad \int_{\mathcal{X}}d\pi(x,y) = d\nu(y).
\end{equation}

The problem can now be posed as the optimization program 
\begin{equation}
    \begin{aligned}
        \min_{\pi \in \mathcal{M}_+(\mathcal{X} \times \mathcal{Y})} \quad &\int_{\mathcal{X}\times\mathcal{Y}}c(x,y)d\pi(x,y) \\ 
        \text{s.t.} \qquad\quad & \int_{y \in \mathcal{Y}}d\pi(x,y) = d\mu(x), \\
        \quad & \int_{x \in \mathcal{X}}d\pi(x,y) = d\nu(y),
    \end{aligned}
\end{equation}
where the cost function $\mathcal{I}(\pi) = \int_{\mathcal{X}\times\mathcal{Y}}c(x,y)d\pi(x,y)$ is the total transportation cost under the transport plan $\pi$. This form is known as Kantorovich's optimal transportation problem, and is a relaxation of the original formulation by Monge. The interested reader is referred to \cite{Villani2003, COTFNT, Chen2021} for more details.

\section{Optimal Transport for Correctional Learning}
In this section we present the main contribution of this paper: an optimal transport formulation of the \textit{batch} correctional learning problem. The main motivation behind using optimal transport is to create a general framework that allows for the estimation of more complex parameters.

\subsection{General Problem Formulation}
Recall the problem setup of correctional learning in Section~\ref{subsec:corr_learning}. We now put it into a more general setting to make its connections to optimal transport more clear.

Assume that the data-generating system is permutation-invariant, or \textit{exchangeable}, in the sense that the distribution $p_0^N$ of the samples $\mathcal{O}_N$ it generates does not change if the samples in $\mathcal{O}_N$ are permuted (in a deterministic manner). It is then natural to consider estimators that are also permutation-invariant, i.e., that can be described as some function of the samples' empirical measure: 
\begin{equation}
    \hat{\theta}_N(x_1, \hdots, x_N) = J(\hat{p}_N(x_1, \hdots, x_N)),
    \label{eq:ot_estimator}
\end{equation}
where $\hat{p}_N\colon \mathcal{X}^N \rightarrow \mathcal{M}_+(\mathcal{X})$ is the empirical measure of the samples, defined as 
\begin{equation}
    (\hat{p}_N (x_1, \hdots, x_N))(A) := \sum_{i=1}^N\frac{1}{N}\mathds{1}\{{x_i = A\}}, \quad A \in \mathcal{B}.
\end{equation}
The function $J\colon \mathcal{M}_+(\mathcal{X}) \to \Theta$ is a fixed function (i.e., independent of $N$), which we will later assume to be Fréchet-differentiable.

Recall that the samples can be perturbed by the teacher before they reach the student. In the batch setting, the teacher has access to all of the original samples, $\mathcal{O}$, before perturbing them into $\Tilde{\mathcal{O}} = (\Tilde{x}_1, \hdots, \Tilde{x}_N )$, where $\Tilde{x}_i \in \Tilde{\mathcal{X}} \subseteq \mathbb{R}^d$ for all $i$. The teacher is subject to a budget constraint, namely
\begin{equation}
    \sum_{i=1}^Nc(x_i, \tilde{x}_i) \leq B.
    \label{eq:ot_budget_constraint}
\end{equation}
The goal of the teacher is still to modify the original sequence, $\mathcal{O}$, subject to \eqref{eq:ot_budget_constraint}, in order to minimize the estimation error 
\begin{equation}
    \lVert \theta_0 - \Tilde{\theta} \rVert,
\end{equation}
where $\Tilde{\theta}$ is the altered estimate based on $\tilde{\mathcal{O}}$.

Since the estimator in \eqref{eq:ot_estimator} depends on the samples only through their empirical distribution, it makes sense to pose the optimization problem to be solved by the teacher in terms of distribution functions, i.e., as an optimal transport problem 
\begin{subequations}
    \begin{align}
    \min_{p} \  &   \left \lVert \theta_0 - J\left(\int_{x\in\mathcal{X}} dp(x, \cdot)\right) \right\rVert ^2 \label{eq:opt_prob2_objective} \\
    \textrm{s.t.} \   &\int_{(x,\Tilde{x})\in \mathcal{X}\times\mathcal{\tilde{X}}}c(x, \Tilde{x})dp(x, \Tilde{x})\leq \frac{B}{N}\label{eq:opt_prob2_budget_constraint} \\
        & 
        \begin{aligned}
            \int_{\tilde{x}\in\tilde{\mathcal{X}}}\int_{x\in A}dp(x, \tilde{x}) &= (\hat{p}_N(x_1,\hdots,x_N))(A), \\
        &\forall { } A\in\mathcal{B},
        \end{aligned}
        \label{eq:opt_prob2_marginal_constraint}
    \end{align} 
    \label{eq:opt_prob2}
\end{subequations}
where $p \in \mathcal{M}_+(\mathcal{X} \times \mathcal{\tilde{X}})$ is a transportation map representing the joint measure of the original and modified samples, and $\hat{p}_N$ the empirical distriubtion of the original samples. We note that this is in general an infinite-dimensional problem with linear constraints. In general, however, $J$ is not necessarily linear. In the case that $J$ is Fréchet-differentiable, and we assume that the budget $B$ is ``small'' (in the sense that most of the original samples will not be modified), one can use the Taylor approximation of the cost of the modified samples, 
\begin{equation}
    \begin{split}
        &J\left(\int_{x\in\mathcal{X}} dp(x, \cdot) \right) 
        \approx J(\hat{p}_{N}) \\
         &+ 
        \int_{x\in\mathcal{X}}\int_{\tilde{x}\in \tilde{\mathcal{X}}} (\nabla J(\hat{p}_{N}))(\tilde{x})dp(x, \tilde{x}) \\
        & -  
        \int_{x\in\mathcal{X}}(\nabla J(\hat{p}_{N}))(x) d\hat{p}_{N}(x),
    \end{split}
    \label{eq:taylor_approx_of_J}
\end{equation}
where $\hat{p}_{N}$ is the empirical distribution of the original sample sequence $\mathcal{O}$. We let $J_{\text{TA}}(\mathcal{O})$ denote the Taylor approximation. Substitution into \eqref{eq:opt_prob2_objective} then yields the objective 
\begin{equation}
    \min_p \ \lVert \theta_0 - J_{\text{TA}}(\mathcal{O}) \rVert ^2.
    \label{eq:ta_objective}
\end{equation}

\begin{remark}
Note how the constraints in \eqref{eq:opt_prob2} now consider the distribution of the samples. This is in contrast to the original formulation in \eqref{eq:cl_batch_opt_prob}, in which each observation is considered individually. 
\end{remark}

\subsection{Discretization of the Continuous Case}
One of the most common approaches to solve the optimal transport problem in \eqref{eq:opt_prob2} is to discretize it \cite{Chen2021}. For simplicity, we will assume that the original samples are independent and identically distributed, with distribution $p_0$. We start by defining a discretized sample space. Recall that our observation sequence is given by the \textit{multiset}\footnote{A sample may occur multiple times in the sequence.} 
\begin{equation}
    \mathcal{O} = \{x_1, \hdots, x_N\}. 
\end{equation}
We can let the \textit{set} of unique values of $\mathcal{O}$ constitute our discretized sample space as 
\begin{equation}
    \mathcal{S} = \bigcup_{o \subseteq \mathcal{O}} o = \{s_1, \hdots, s_n\} \subseteq \mathcal{O}.
\end{equation}
For the continuous case, we note that with probability one,
\begin{equation}
    \mathcal{S} = \mathcal{O} \quad \text{and} \quad n = |\mathcal{S}| = |\mathcal{O}| = N,
\end{equation}
since all the samples in $\mathcal{O}$ are distinct, with probability one. The elements in $\mathcal{S}$ will be called \emph{states}.
\begin{remark}
We note that there are other methods to determine the states. For instance, they may be fixed to belong to some pre-determined set of values. However, with regards to the nature of the framework of modifying an \textit{observed} sequence, we believe the suggested approach is reasonable.
\end{remark}

Furthermore, the teacher will be allowed to change the observations in $\mathcal{O}$ into samples from the set 
\begin{equation}
    \Tilde{\mathcal{S}} = \{\tilde{\bm{s}}_1, \hdots, \Tilde{\bm{s}}_m\}, 
\end{equation}
which may or may not coincide with $\mathcal{S}$. Note that $|\Tilde{\mathcal{S}}| = m$.

We now continue by discretizing \eqref{eq:opt_prob2}.
We note that both the objective in \eqref{eq:taylor_approx_of_J} as well as our constraints in \eqref{eq:opt_prob2} include our decision variable $dp(x, \Tilde{x})$ in the integrals, which we cannot sample from. Thus, to approximate the integrals, we use techniques from importance sampling \cite{TheodoridisMLBook}. We consider the proposal distribution
\begin{equation}
    d\mu(x,\tilde{x}) = dq(x)dr(\tilde{x}),
\end{equation}
where $dq(x)$ and $dr(\tilde{x})$ are probability measures defined on $\mathcal{S}$ and $\Tilde{\mathcal{S}}$, respectively. Note that, for simplicity, $\mu$ has been chosen in terms of independent proposal distributions for $x$ and $\Tilde{x}$. With this distribution, and restricting $p$ to $\mathcal{S} \times \Tilde{\mathcal{S}}$, we rewrite the estimator in \eqref{eq:taylor_approx_of_J} as
\begin{equation}
\begin{aligned}
        &J_{\text{TA}}(\mathcal{O}) = J(\hat{p}_{N}) 
        \\ 
        & + 
        \int_{x\in\mathcal{S}}\int_{\tilde{x}\in\tilde{\mathcal{S}}} (\nabla J(\hat{p}_{N}))(\tilde{x})\frac{dp(x, \tilde{x})}{dq(x)dr(\tilde{x})} dq(x)dr(\tilde{x}) \\
        & -  
        \int_{x\in\mathcal{S}}(\nabla J(\hat{p}_{N}))(x) d\hat{p}_{N}(x).
    \end{aligned}
    \label{eq:taylor_expansion_2}
\end{equation}
Using importance sampling, we can discretize the expression in \eqref{eq:taylor_expansion_2} as 
\begin{equation}
    \begin{aligned}
        &J_{\text{TAD}}(\mathcal{O}) = 
        J(\hat{p}_{N}) \\
        & \quad\qquad + 
        \frac{1}{nm} \sum_{i=1}^n\sum_{j=1}^m \frac{\partial J}{\partial p_{\Tilde{x}_j}}(\hat{p}_{N}) \frac{dp(x_i,\Tilde{x}_j)}{dq(x_i)dr(\Tilde{x}_j)} \\
        & \quad\qquad - 
        \frac{1}{m}\sum_{j=1}^m\frac{\partial J}{\partial \Tilde{x}_j}(\hat{p}_{N}),
    \end{aligned}
    \label{eq:ta_disc}
\end{equation}
where we use numerical differentiation to approximate the gradient $\nabla J$. To simplify the notation, we define 
\begin{equation}
    \alpha \in \mathbb{R}^{n\times m}: \ \alpha_{ij} = \frac{dp(x_i, \Tilde{x}_j)}{dq(x_i)dr(\Tilde{x}_j)}
    \label{eq:alpha_definition}
\end{equation}
to be our new decision variable. Our objective in \eqref{eq:ta_objective} can then be written as 
\begin{equation}
    \min_\alpha \ \lVert \theta_0 - J_{\text{TAD}}(\mathcal{O}) \rVert ^2.
    \label{eq:objective_disc}
\end{equation}
We discretize the budget constraint \eqref{eq:opt_prob2_budget_constraint} as 
\begin{equation}
    \frac{1}{nm}\sum_{i=1}^n\sum_{j=1}^m c(x_i, \tilde{x}_j)\alpha_{ij} \leq \frac{B}{N}.
    \label{eq:disc_budget_constraint}
\end{equation}
To discretize the constraint in \eqref{eq:opt_prob2_marginal_constraint}, we first utilize the same trick as before and rewrite it as 
\begin{equation}
        \int_{x\in\mathcal{S}} \frac{dp(x, \Tilde{x})}{dq(x)dr(\Tilde{x})}dr(\Tilde{x}) = \frac{d\hat{p}_N(x)}{dq(x)},
\end{equation}
We discretize this as 
\begin{equation}
    \frac{1}{m}\sum_{j=1}^m \frac{dp(x,y_j)}{dq(x_i)dr(y_j)} = \frac{d\hat{p}_N(x)}{dq(x_i)}.
\end{equation}
Again, since we cannot sample from $p(x, \Tilde{x})$, we simply say that the above relation must hold for all values of $x$, i.e., 
\begin{equation}
    \frac{1}{m}\sum_{j=1}^m \underbrace{\frac{dp(x_i,\Tilde{x}_j)}{dq(x_i)dr(\Tilde{x}_j)}}_{\alpha_{ij}} = \frac{d\hat{p}_N(x_i)}{dq(x_i)}, \quad \forall i.
    \label{eq:marginal_constraint_discretized}
\end{equation}
All constraints are now written in terms of our new decision variables $\alpha_{ij}$.

\subsection{Importance Sampling: Different Approaches}
Consider the case when $\Tilde{\mathcal{S}} = \mathcal{S}$, and $dq = dr = \hat{p}_N$. This means that we we will work directly with the observed samples, and the constraint in \eqref{eq:marginal_constraint_discretized} then simplifies to 
\begin{equation}
    \frac{1}{m}\sum_{j=1}^m\frac{dp(x_i,y_j)}{dq(x_i)dr(x_j)} = \frac{d\hat{p}_N}{dq}(x_i) = \frac{d\hat{p}_N}{d\hat{p}_N}(x_i) = 1\quad \forall i.
\end{equation}
We note that this case is very similar to a discrete setting in the sense that we are limiting the teacher to change the observations into values that have already been seen or encountered. This approach is similar to other resampling techniques and can be viewed as a way of re-weighting the samples to change their importance for the estimation. 

We can also sample from $dq$ and $dr$ independently, with $dq \neq dr$. Using this approach, we can impose some prior knowledge on $dr$, either by defining it to be the true distribution, or some distribution that will yield a more accurate estimate of the parameter we are interested in. However, using this approach, we would not be working directly with the empirical distribution, which means that we would have to perform an interpolation step to figure out how to best change the actual observations. We would also have to use a density estimation technique to enforce the constraint in \eqref{eq:opt_prob2_marginal_constraint}.

\subsection{Modifying the Sequence}
Next we describe how the teacher modifies the sequence based on the $\alpha$ obtained from solving \eqref{eq:objective_disc} w.r.t.\ the constraints \eqref{eq:disc_budget_constraint} and \eqref{eq:marginal_constraint_discretized}. Recall the definition of $\alpha$ in \eqref{eq:alpha_definition}, and that $dp(x, \Tilde{x})$ denotes the amount of probability mass transferred from $x$ to $\Tilde{x}$. Then, by applying Bayes' theorem, we compute the conditional probability mass function (pmf) as 
\begin{equation}
    p(\Tilde{x} \mid x) = \alpha \hat{p}_N(\Tilde{x}),
    \label{eq:conditionaL_pmf}
\end{equation}
which will give us the probability of changing an observation $x$ into $\Tilde{x}$. 

The teacher's intervention procedure is then as follows. For each sample in $x_i \in \mathcal{O}$, the teacher modifies it according to the pmf in \eqref{eq:conditionaL_pmf}. That is, 
\begin{equation}
    x_i \ \rightarrow \  \Tilde{x}_i, \ \Tilde{x}_i \sim  p(\Tilde{x} \mid x_i).
\end{equation}

To ensure that the intervention budget is not exceeded, the teacher generates $M$ new sequences. For each new generated sequence that satisfy the budget constraint (i.e., for which the number of corrections is less than or equal to $B$), an updated estimate is computed. Out of these sequences, the one yielding the lowest estimation error is then chosen to be the optimal one. Should the teacher fail to find a sequence that both improves the estimate and satisfies the budget constraint, it will keep the original sequence.

Naturally, the teacher may follow different intervention policies. Alternative approaches may include changing one sample at a time and then re-solve for a new $\alpha$ following each update. This policy is similar to receding horizon control strategies where we may interpret the budget to be the horizon \cite{borrelliMPC}.

Another possible strategy would be for the teacher to always make the change with the highest probability. This would make for a greedy approach \cite{CormenAlgorithms}.

\section{Numerical results}
In this section, we evaluate our framework in three different settings; two theoretical and one applied. For simplicity, we consider $\mathcal{X} \in \mathbb{R}$ and $\theta_0 \in \mathbb{R}$ in all settings. We evaluate the performance in terms of the absolute error, i.e., $\lvert e \rvert = \lvert \theta_0 - \Tilde{\theta} \rvert$. 

\subsection{Variance Estimation of a Gaussian Distribution}
In the first experiment, we use the framework to estimate the variance of a Gaussian distribution. We consider the observations to be sampled from the distribution $\mathcal{N}(0,1)$, so $\theta_0 = \sigma^2 = 1$.

\begin{figure}
    \centering
    \includegraphics[width=0.35\textwidth]{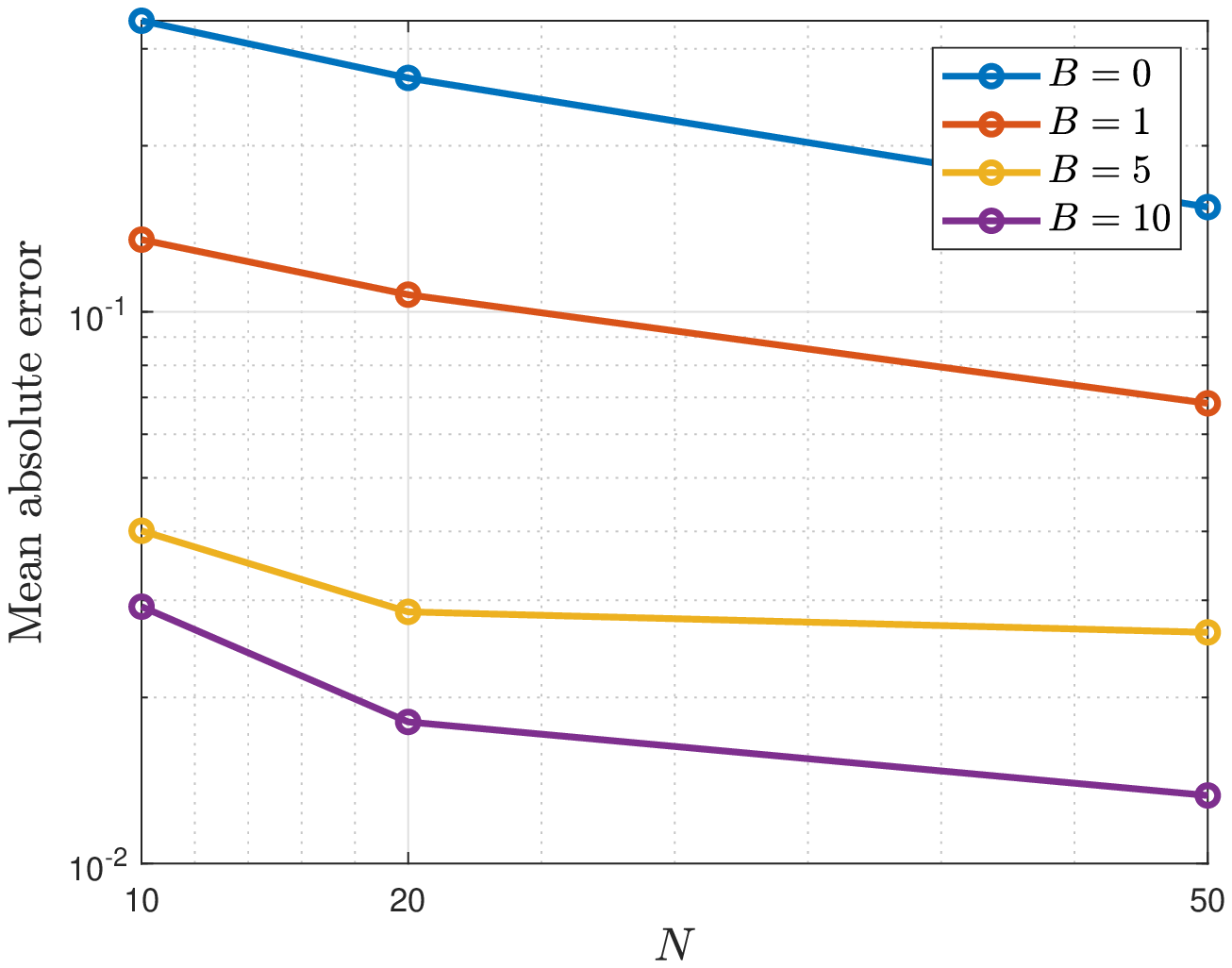}
    \caption{The absolute estimation error averaged over 100 Monte Carlo simulations for increasing sample sizes and budgets. Note that $b=0$ corresponds to the case with no teacher intervention.}
    \label{fig:mc_variance_estimation_gaussian}
\end{figure}

We perform the estimation on three sample sizes, $N = \{10, 20, 50\}$, subject to four different intervention budgets, $B = \{0, 1, 5, 10\}$. In this example, we use a uniform transportation cost, i.e., $c(x, \tilde{x}) = \mathbb{I}$. This means that all changes made are equally expensive. For each sample size and budget, we perform the experiment 100 times and compute the average absolute error. For all configurations, we use $M = 1000$. 
The results are shown in Figure \ref{fig:mc_variance_estimation_gaussian}. As expected, the plot shows a decrease in the estimation error as the sample size increase. It also shows that the error is further decreased as the budget increases.

\subsection{Scale Estimation of a Weibull Distribution}
Next we apply the framework to estimate the scale parameter of a Weibull distribution. The probability density function is given by 
\begin{equation}
    f(x) = 
    \begin{cases}
    \begin{aligned}
        &\frac{k}{\lambda}\left(\frac{x}{\lambda}\right)^{k-1}e^{-(x-\lambda)^k}, \quad &&x \geq 0,\\
        &0, \quad &&x < 0,
    \end{aligned}
    \end{cases}
\end{equation}
where $k > 0$ is called the shape parameter, and $\lambda > 0$ the scale parameter. In this example, we will consider the estimation of $\lambda$ of a Weibull distribution with $\theta_0 = \lambda_0 = 2$ and $k = 8$.

\begin{figure}
    \centering
    \includegraphics[width=0.35\textwidth]{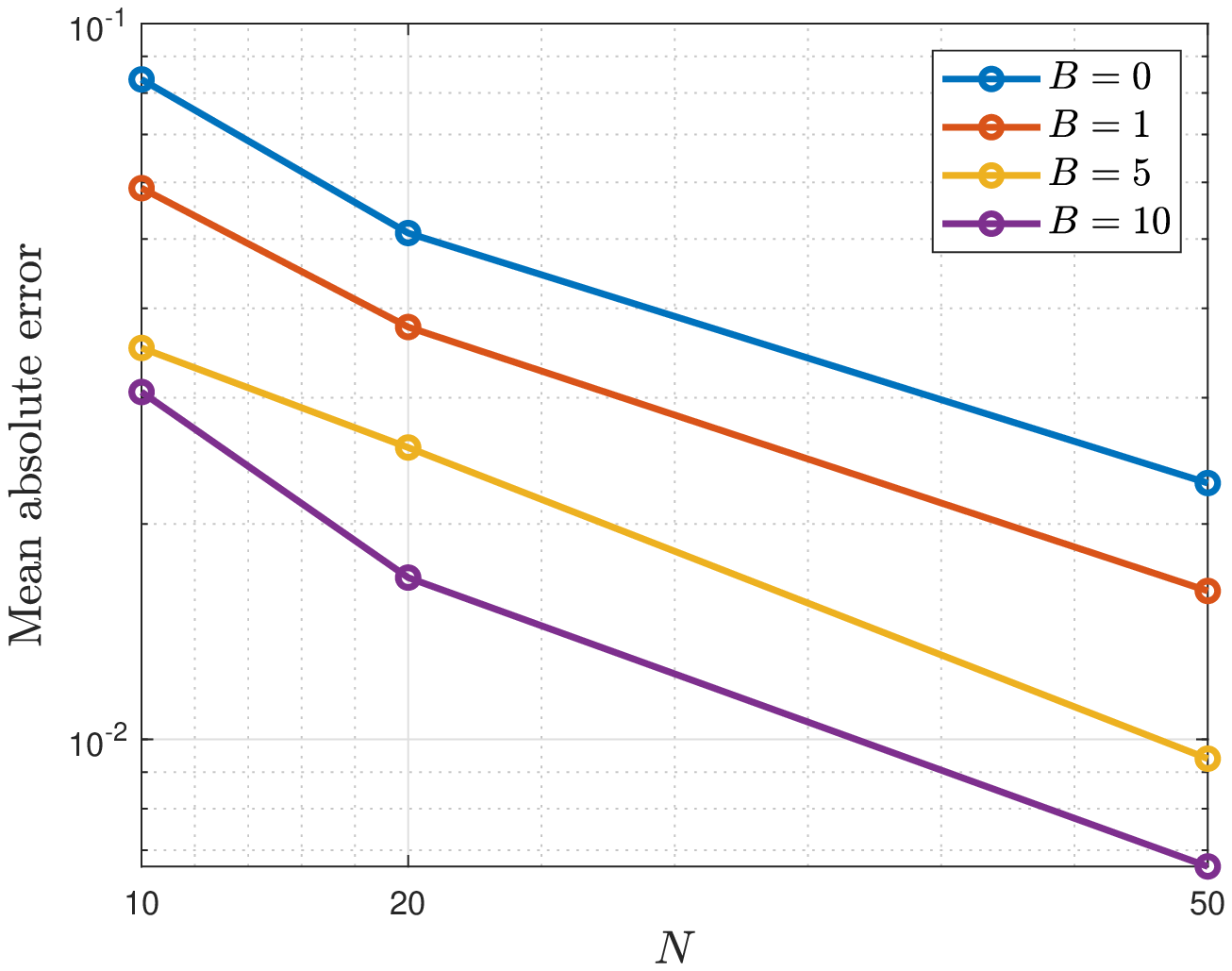}
    \caption{The absolute estimation error averaged over 100 Monte Carlo simulations for increasing sample sizes and budgets. Note that $b=0$ corresponds to the case with no teacher intervention.}
    \label{fig:mc_scale_estimation_weibull}
\end{figure}

There are different approaches available for estimating $\lambda$, see e.g.\cite{weibull_estimation}. In this experiment, we use the Bayesian two-stage approach derived in \cite{Lakshminarayanan2022}. We use the proportional cost 
\begin{equation}
    c(x, \Tilde{x}) = 10 \times \lceil \lvert \Tilde{x} - x\rvert \rceil, 
\end{equation}
where $\lceil \cdot \rceil$ denotes the ceiling function. As in the previous example, we run the estimation process on the sample sizes $N = \{10,20,50\}$  and for the intervention budgets $B = \{0,1,5,10\}$. We use $M=2000$ for all configurations. The averaged estimation errors are shown in Figure 
\ref{fig:mc_scale_estimation_weibull}. The results are similar to what we observed in the previous experiment, with an improved estimation error for increasing sample sizes and budgets.

\subsection{Reward Estimation in Inverse Reinforcement Learning}
As a final example, we apply our framework to update a robot's reward function in an inverse reinforcement learning setting. Recent work on learning from human interaction shows how physical corrections made by a human (e.g.\ in the form of applied torque) can improve a robot's learning process \cite{bajcsy2017learning}. Inspired by their problem setup, we apply our framework in a similar setting. 

\begin{figure}[t]
 \centering
    \includegraphics[width=0.35\textwidth]{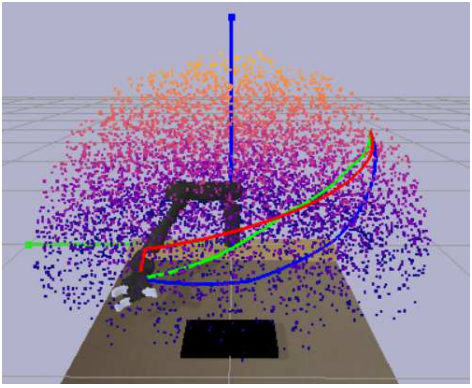}
    \caption{The robot observes $N=3$ trajectories with different feature values. The expert may alter some of them to the one that is closer to its preferences. For example, if the robot should avoid the laptop, the expert may change the blue trajectory into the red one, to reflect this.}
    \label{fig:robot_reward}
\end{figure}

Consider a robot arm being tasked with moving a coffee cup from one side of a table to the other. To learn the task, the robot gets to observe a set of $N$ trajectories, $\{\xi_1, \hdots, \xi_N\}$, demonstrated by a human. Figure \ref{fig:robot_reward} illustrates some examples of trajectories demonstrated on a robotic arm with seven degrees of freedom implemented in PyBullet. Each trajectory is associated with a total feature count for each feature $i \in [n]$
\begin{equation}
    \Phi_i(\xi) = \sum_{x \in \xi}\phi_i(x),
\end{equation}
where $\phi_i(x)$ is the local feature value in a point $x$ along the trajectory $\xi$. A high feature value corresponds to a good position in space. The features represent different subgoals in performing the task, such as ``stay nearby the top of the table'' and ``avoid the laptop''. Based on these features, the robot learns a reward function 
\begin{equation}
    R = \Theta^T \Phi = \theta_1 \Phi_1 + \hdots \theta_n \Phi_n,
\end{equation}
where the weights $\Theta$ represent the importance of each feature to the human.

Assume now that the robot has learned a reward function based on the following observations collected over $N = 5$ trajectories with $n=3$ features 
\begin{equation}
    \begin{cases}
        \bm{\Phi_1} = \{\Phi_1(\xi_i)\}_{i=1}^5 =  \{100, \ 75,\ 50, \ 20, \ 5\} \\
        \bm{\Phi_2} = \{\Phi_2(\xi_i)\}_{i=1}^5 = \{90,\ 200, \ 10, \ 2, \ 30\} \\
        \bm{\Phi_3} = \{\Phi_3(\xi_i)\}_{i=1}^5 = \{50, \ 20, \ 3, \ 5, \ 10\}
    \end{cases}.
    \label{eq:robot_observations}
\end{equation}

An expert may then apply our framework to improve the robot's learned $\theta_i$'s, by modifying the sets $\bm{\Phi}_i$ in \eqref{eq:robot_observations} into $\Tilde{\bm{\Phi}}_i$. As estimator, we consider a slightly modified version of the weight update in \cite{bajcsy2017learning}:
\begin{equation}
    \Tilde{\theta}_i = \hat{\theta}_i + \beta \left ( \sum_{\Phi \in \bm{\Phi}} \Phi - \sum_{\tilde{\Phi} \in \Tilde{\bm{\Phi}}} \tilde{\Phi} \right ),
\end{equation}
where $\beta < 0$ is a step/scaling parameter. Here, we consider $\beta = -0.001$. Note how the feature weights are updated based on the direction of change of the feature values between the original and the modified trajectories. If the altered corrections pass further away from, say, the laptop, the $\theta_i$ corresponding to the distance-to-laptop feature will increase.

For this experiment we used $c(x, \Tilde{x}) = \mathbb{I}$, $b=1$, and $M=1000$. The corrected feature values together with their corresponding updated weight estimate are shown in Table \ref{tab:robot_results}. For reference, we also we also present the true weights and the initial estimates in the same table. The results show that the updated estimates are closer to the true values, compared to the initial estimates.

\begin{table}
\caption{The corrected feature values and their corresponding estimates.}
	\begin{center}
		\begin{tabular}{l|l|l|l}
            True weight & Old est. & New est. & Corrected feature values \\
			\hline 
			$\theta_1 = 0.1$ & $\hat{\theta}_1 = 0.5$ & $\Tilde{\theta}_1 = 0.05$ & $\Tilde{\bm{\Phi}}_1 = \{100, \ 75, \ 50,\ 20,\ \textbf{50} \}$ \\
			$\theta_2 = 1$   & $\hat{\theta}_2 = 0.5$ & $\Tilde{\theta}_2 = 1.1$  & $\Tilde{\bm{\Phi}}_2 = \{\textbf{30},\ 200,\ 10,\ 2,\ 30 \}$ \\
			$\theta_3 = 0.8$ & $\hat{\theta}_1 = 0.5$ & $\Tilde{\theta}_3 = 0.8$  & $\Tilde{\bm{\Phi}}_3 = \{\textbf{20},\ 20,\ 3,\ 5,\ 10 \} $\\
		\end{tabular}
	\end{center}
\label{tab:robot_results}
\end{table}


\section{CONCLUSION AND FUTURE WORK}
In this work, we presented a generalized formulation of the correctional learning framework using optimal transport. We demonstrated that by expressing the correctional learning problem as an optimization program in terms of distribution functions, we obtain a more general and flexible framework better suited for estimation of more complex characteristics. We successfully applied the framework on three estimation processes; for the variance estimation of a Gaussian, the scale estimation of a Weibull, and, finally, in an inverse reinforcement setting where we improved a robot's reward function. 
This novel optimal-transport formulation opens up for several interesting extensions, including using correctional learning for differential privacy, considering time-series data, and for balancing biased or skewed learning datasets. Addressing the curse of dimensionality of the discretized problem will be an important step along the way.

\section*{Acknowledgement}
The authors would like to thank Isabel Haasler for valuable discussions on Optimal Transport theory.


\bibliographystyle{IEEEtran}
\bibliography{references}

\end{document}